\documentclass[12pt]{article}
\usepackage{times} 
\usepackage{url} 
\usepackage{graphicx} 
\usepackage[margin=0.8in]{geometry} 
\usepackage{authblk} 
\usepackage{amsmath} 
\usepackage{amsfonts}
\usepackage{tabularx,ragged2e,booktabs} 
\usepackage{titlesec}
\usepackage{amssymb}
\usepackage{comment}
\usepackage{subcaption}
\usepackage{authblk}
\usepackage{xcolor} 
\usepackage{hyperref}
\usepackage{amssymb}
\usepackage{dsfont}
\usepackage{float}
\usepackage{multirow}
\usepackage[normalem]{ulem}
\usepackage{float}
\usepackage{graphics}
\usepackage{multirow}
\usepackage{graphicx,lipsum,wrapfig}%
\usepackage[labelfont=bf,labelsep=period,justification=justified]{caption}
\usepackage{lineno}
\usepackage[nobiblatex]{xurl}
\usepackage[subtle]{savetrees}
\usepackage{algorithm} 
\usepackage{algorithmic}  
\usepackage{bbm}
\usepackage{dsfont}
\usepackage{newtxmath}

\def\cond{\, | \,}

\author[1, *]{Whitney Sloneker}
\author[2]{Shalin Patel}
\author[1]{Michael Wang}
\author[1, 3]{Lorin Crawford}
\author[1, 2]{Ritambhara Singh}

\affil[1]{Center for Computational Molecular Biology, Brown University, Providence, Rhode Island, USA}
\affil[2]{Department of Computer Science, Brown University, Providence, Rhode Island, USA}
\affil[3]{Microsoft Research, Cambridge, Massachusets, USA}
\affil[*]{whitney\_sloneker@brown.edu, ORCID: 0000-0002-5955-1507}
\date{\vspace{-1.5cm}}

\title{BetaExplainer: A Probabilistic Method to Explain Graph Neural Networks}

\begin{document}

\maketitle              

\begin{abstract}
\noindent
Graph neural networks (GNNs) are powerful tools for conducting inference on graph data but are often seen as "black boxes" due to difficulty in extracting meaningful subnetworks driving predictive performance. Many interpretable GNN methods exist, but they cannot quantify uncertainty in edge weights and suffer in predictive accuracy when applied to challenging graph structures. In this work, we proposed BetaExplainer which addresses these issues by using a sparsity-inducing prior to mask unimportant edges during model training. To evaluate our approach, we examine various simulated data sets with diverse real-world characteristics. Not only does this implementation provide a notion of edge importance uncertainty, it also improves upon evaluation metrics for challenging datasets compared to state-of-the art explainer methods.\\


\end{abstract}
\section{Introduction}
Relational data occur in a variety of domains, such as social graphs \cite{GNNBenefits}, chemical structures \cite{SurveyGNN}, physical systems \cite{GNNBenefits}, gene-gene interactions \cite{GNNBenefits}, and epidemiological modeling \cite{EpModel}. These data are best represented by graphs that effectively model their relationships, such as chemical bonds in drug molecules that affect toxicity or treatment efficacy \cite{GNNBenefits} or personal interactions in social networks indicating contact \cite{SurveyGNN}. Although graph information represents these datasets more accurately by incorporating node features (i.e., chemical weight for molecules) and node interactions through edges (i.e., chemical bonds) \cite{GNNBenefits}, large-scale modeling to learn their patterns can be challenging if the graphs are complex \cite{GNNRealWorld, TrustworthyGNN}. 
Embedding methods such as Graphlets\cite{Graphlets} and DeepWalk\cite{DeepWalk} have been developed to address these challenges. However, they may oversimplify complex graphical features by summarizing the graph and ignoring node features \cite{GraphEmbed}. As a result,  graph neural networks (GNNs) have been widely adopted in the machine learning community to model graph-based datasets because they incorporate edge structure and node features directly \cite{GNNTax, GNNParadigm}.

GNN models have broad applications, such as capturing the complexity of traffic dynamics, approximating NP-hard graph combinatorial analysis, and learning real-world graphs such as molecule structures \cite{GNNBenefits}. However, like other deep learning models, they can be hard to explain \cite{GNNTrust}. It is challenging to extract important edges from which the GNN is learning to make accurate predictions \cite{GNNTrust}. Determining these important edges is needed for hypothesis development \cite{GNNTrust}, such as answering "what chemical bonds might determine the prediction of toxicity in a molecule?". As a result, it is critical to develop GNN explanation methods to understand the model predictions that highlight the graph's important edges.

Many methods have been created to explain GNNs, that is, highlight the important edges for predictions, but their performances vary widely depending on the underlying properties of the data or the GNN model. For example, gradient-based explainability methods struggle to produce accurate edge explanations when deep learning models experience gradient saturation \cite{Explain}. Data properties can also influence the explainer's performance. A recent benchmark study \cite{GraphXAI} found that many methods (Grad \cite{Grad}, GradCAM \cite{GradCam}, GuidedBP \cite{GuidedBP}, Integrated Gradients \cite{IntegratedGradients}, GNNExplainer \cite{GNNExp}, PGExplainer \cite{PGExp}, SubgraphX \cite{SubgraphX}, and PGMExplainer \cite{PGMExp}) struggle against challenging data properties. For example, if the underlying graph of the data is heterophilic\footnote{When a graph is highly heterophilic, it means edges tend to connect nodes of different classes. Its opposite, a homophilic graph, means edges tend to connect nodes of the same class.} or a low proportion of node features are critical for classification, the existing explainer methods struggle to produce accurate edge explanations from the GNN model. 

The study shows that out of all the existing methods, only GNNExplainer, PGExplainer, and SubgraphX function as effective edge explainers for GNN models \cite{GraphXAI}. However, PGExplainer often underperforms most methods in generating accurate edge explanations for the simulated datasets \cite{GraphXAI}. On the other hand, SubgraphX has a robust performance \cite{GraphXAI}. However, it cannot rank the edge importance as each highlighted edge is only denoted as 0 (as unimportant) or 1 (as important) \cite{Explain}. This ranking is essential to hypothesis generation as it allows researchers to focus on the most highly ranked edges, thus easing downstream analysis. For instance, in the case of exploring edges representing gene-gene interactions for biological datasets, edges ranked based on their importance allow us to focus on experimentally confirming only the most likely interactions, saving the time and monetary cost required for wet lab experiments \cite{GCNLongRange}. Finally, GNNExplainer performs relatively well on the proposed synthetic datasets \cite{GraphXAI} while also returning importance scores that can be used to rank the relevant edges of the GNN model \cite{Explain}. As a result, GNNExplainer best represents state-of-the-art edge explainers for GNNs. Even so, its struggle to produce accurate edge explanations suggests a knowledge gap for explainer methods on the tested challenging datasets. 

We explored another challenging setting for GNN explainer methods -- explaining a GNN model for graph datasets constructed with sparse node features. A dataset with sparse node features is one where many node features are zeros. Sparse node feature datasets are common in various real-world domains, especially in the now-emerging single-cell gene expression data (or scRNA-seq). These datasets are notoriously sparse, and any gene-gene correlation or interaction graphs created from them will result in graphs with low informative node features \cite{SparseRNA} \cite{GCNLongRange}. We hypothesize that existing explainers will also struggle to accurately return edge explanations for the sparse node feature dataset. However, due to their commonality, thus, ensuring explainability methods succeed on sparse node datasets is necessary.

\begin{figure}[H]
\centering
\includegraphics[scale=0.5]{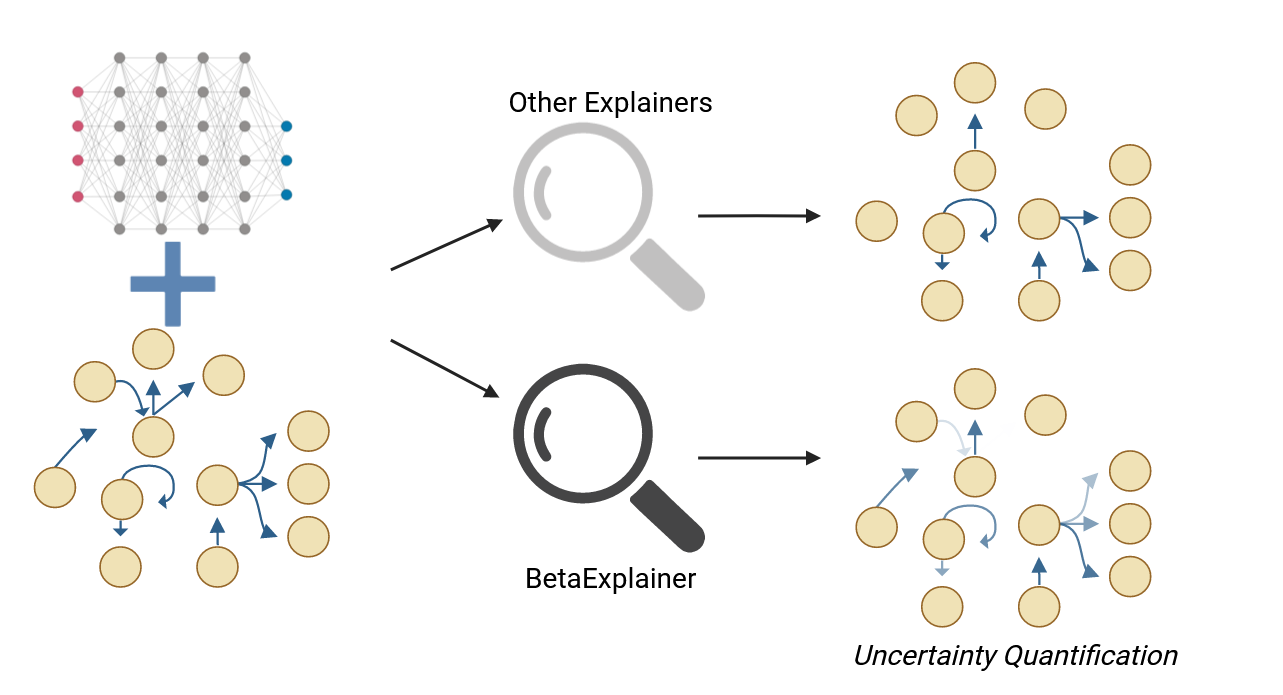} 
\caption{BetaExplainer returns a mask for the important edges of a graph for a GNN's classification. As it learns a probabilistic model to represent these important edges, the mask contains a level of uncertainty quantification to indicate the importance of each edge to a GNN missing or limited in other models}
\label{fig:Fig1}
\end{figure}

We propose a new method, BetaExplainer, that uses a probabilistic distribution to determine the important edges from a GNN model. BetaExplainer learns a probabilistic edge mask to maximize the similarity of the output of the trained GNN on a masked graph to its original output through statistical inference to approximate which edges are most important (Fig. \ref{fig:Fig1}). This probabilistic approach allows us to produce edge importance scores with uncertainty quantification and rank edges by the order of score confidence. have evaluated BetaExplainer on seven simulated datasets with various challenging underlying data properties that explainers struggle to adapt to, including a heterophilic graph and a sparse node feature dataset. The results demonstrate that BetaExplainer significantly outperforms existing methods in being faithful to the underlying graph importance on five out of seven simulated datasets and improves accuracy compared to state-of-the-art methods on graph datasets with sparse features. It also conveys a notion of uncertainty in edge importance for its explanations, which provides guidance to downstream analysis.

\section{Methods}
\subsection{BetaExplainer Algorithmic Framework}
\begin{figure}[H]
\centering
\includegraphics[scale=0.75]{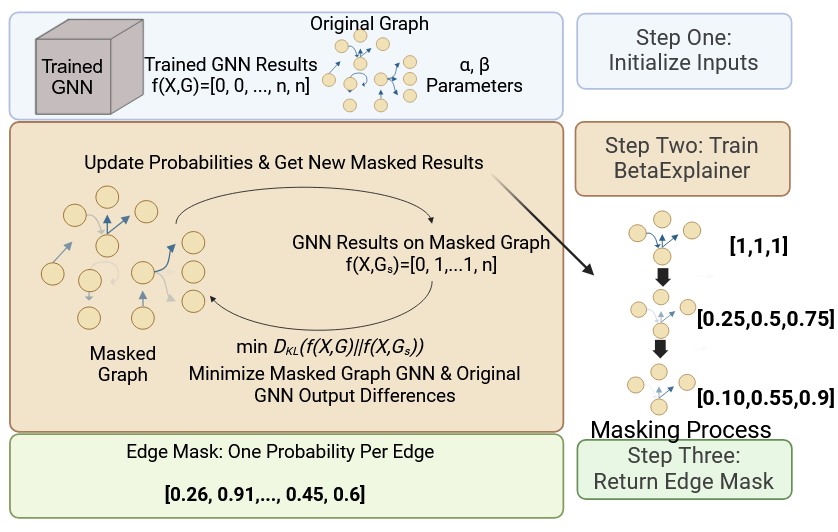}
\caption{Given a trained GNN, the original graph, and Beta distribution parameters $\alpha$ and $\beta$, BetaExplainer is trained by learning the masked-out graph minimizing the KL-Divergence Loss between the model output on the masked out graph and original graph. It will return the learned edge mask representing a probabilistic importance score for each edge when complete}
\label{fig:Fig2}
\end{figure}

Given a trained GNN model $f$,  graph input $G = (V, E)$ with the set of nodes or vertices defined as $V$ and edges $E$ with edge $e_{ij}$ connecting vertices $v_i$ and $v_j$, node features $X$, and model output on the input graph and node features $f(X, G)$, we define a Beta distribution prior $P(M)$ on the edge mask $M$ of the input graph. BetaExplainer learns the posterior Beta distribution $P(M\cond f(X, E))$ of the edge mask by comparing the results of the masked-out graph $G_s$ and a full set of node features or $f(X, G_{s})$  to the original output on the unmasked graph $G$ and all node features or $f(X, G)$ (Fig. \ref{fig:Fig2}) where the likelihood over the GNN or $p(f(X, E))$ is a Bernoulli distribution described by  

\begin{align}
  P(f(X,E) \cond M) =
  \begin{cases}
    p     & \text{if $M \geq 0.5$}, \\
    1 - p & \text{if $M < 0.5$}.
  \end{cases}
\end{align}

Based on the Kullback–Leibler (KL) divergence between the new and original outputs, BetaExplainer updates the edge mask probabilities to increase or decrease each edge importance value in the edge mask as applicable. We optimize evidence lower bound (ELBO) to learn the final edge mask. This edge mask conveys the importance of each edge as a probabilistic importance score.

BetaExplainer has two major benefits: (1) using a probabilistic framing allows us to convey edge importance for easy interpretation and edge rankings while also conveying uncertainty in edge importance, and (2) users may choose distributional parameters most relevant to the underlying data to improve performance by better representing the underlying distribution of edge importance. For this work, we choose the Beta distribution, which is described by
\begin{align}
P(M_{ij} \cond \alpha, \beta) &= \frac{M_{ij}^{\alpha - 1}(1-M_{ij})^{\beta - 1}\Gamma(\alpha + \beta)}{\Gamma(\alpha)\Gamma(\beta)}
\end{align}
where $\alpha$ and $\beta$ denote are real valued shape parameters. Here, $P(M_{ij} \cond \alpha, \beta)$ is the probability that the mask importance for edge $e_{ij}$ is value $M_{ij}$. As edges of a graph can mostly be described by a Bernoulli distribution measuring binary outcomes - indicating importance or unimportance - the Beta distribution functions as the conjugate prior \footnote{When the prior and posterior distributions come from the same family, it is called a conjugate prior.} of the Bernoulli making it a reasonable choice to describe edge importance uncertainty. Equivalently, if the prior edge importance described by the edge mask $P(M)$ is a Beta distribution and the distribution $P(A\cond M)$, indicating whether each edge is important given the Bernoulli distribution, denotes the mask, then the posterior distribution of edge importance $P(M\cond A)$ will also be a Beta distribution.

We assume that the learned Beta distributions for each edge $e \in E$ are independent of each other. Thus, the learned distribution of edge importance for any edge $e_{ij} \in E$ is equivalent to its learned importance for the full mask $M$ through mean field variational inference. Therefore, we may learn the importance of all edges simultaneously. Thus, the ELBO may be calculated as follows with respect to the original output:
\begin{align}
\mathbb{E}[\log(P(G_s, f(X, G)|M)-\log(P(f(X,G)))]
\end{align}
The ELBO is the lower bound for $P(G_s, f(X, G)|M)$, or
\begin{align}
P(G_s, f(X, G)|M) \geq ELBO
\end{align}
This indicates that the difference between the two expressions can be no less than zero and by maximizing the ELBO, the following holds:
\begin{align}
\log(P(G_s, f(X, G)|M) - ELBO = KL(\log(P(f(X,G)))\,\|\,P(G_s, f(X, G)|M))
\end{align} 
Maximizing the ELBO indirectly minimizes the KL divergence between the model output on the masked edges and the original output, learning the optimal mask. While we could calculate the closed form of the Bayes theorem directly in this case, we chose to approximate the true distribution with a variational family instead to increase BetaExplainer's applicability to large-scale real-world datasets. Thus, BetaExplainer's algorithm is formulated as follows:

\begin{algorithm}[H]
\caption{BetaExplainer Model Set Up}\label{alg:cap}
\begin{algorithmic}

\REQUIRE $X, G = (V, E), e_{ij} \in E, \alpha > 0, \beta > 0$ \hfill\COMMENT{Require positive Beta distribution parameters and vertices in the graph}
    
\REQUIRE $c \sim f(X, G)$ \hfill\COMMENT{Return model outputs given the input data.}

$Z \leftarrow z$ \hfill\COMMENT{Define number of training epochs $Z$ as positive integer $z$}

$T \leftarrow 1$ \hfill\COMMENT{Initialize epoch tracker $T$ for training}
\WHILE{$T \leq Z$}
    \item $\alpha \leftarrow \hat{\alpha}$ , $\beta \leftarrow \hat{\beta}$ \hfill\COMMENT{Update parameters $\alpha$ and $\beta$ on the original graph based on the results of the prior iteration.}
    
    \item $M_{ij} \sim Beta(\alpha, \beta)\, \forall e_{ij} \in E$\hfill\COMMENT{Determine edge weights from Beta distribution}
    
    \item $y \leftarrow f(X, G, M)$ \hfill\COMMENT{Return the model output on input dataset with generated weights for edges}
    
    \item $c \sim y$ \hfill\COMMENT{Compare outputs on weighted dataset to original with KL divergence loss to update parameters}

    \item $T \leftarrow T + 1$ \hfill\COMMENT{Update epoch tracker.}
\ENDWHILE

\RETURN{$M_{ij} \forall e_{ij}$} \hfill\COMMENT{Return weights representing edge probabilities over all edges}
\end{algorithmic}
\end{algorithm}

\noindent BetaExplainer uses the \verb|Pyro| framework for variational inference to develop the edge importance model, and the \verb|pytorch_geometric| framework to train all of the GNN models used.

\subsection{Baselines}
We compare BetaExplainer to state-of-the-art methods GNNExplainer \cite{GNNExp} and SubgraphX \cite{SubgraphX} to demonstrate the improved performance of our approach. GNNExplainer performs well on various datasets for both node and graph classification tasks, suggesting it is a strong baseline \cite{Explain}. Furthermore, while BetaExplainer and GNNExplainer have similar optimization goals, the different training algorithms ensure that BetaExplainer has certain beneficial properties. GNNExplainer randomly initializes the edge mask and then directly optimizes the Bernoulli distribution of edge importance, requiring a re-parameterization trick. In contrast, BetaExplainer directly shrinks edges by using the black-box variational inference. Furthermore, GNNExplainer does not directly drop out parameters by forcing them to be zero; based on the distribution learned, BetaExplainer does this as needed. Using the BetaExplainer method gives us access to different properties, such as incorporating prior information through hyperparameters. This allows the proposed method to easily adapt to challenging data properties such as a heterophilic graph or high sparse node features. Furthermore, as BetaExplainer directly learns a distribution, it can convey a notion of uncertainty in edge importance. 

SubgraphX uses Markov Chain Tree Search to determine the subgraph that achieves a Shapley value, suggesting that the model results on the subgraph are similar to the original results. While this is similar to BetaExplainer - both attempt to learn the subgraph that returns similar model results to the initial - it does not allow for a notion of uncertainty like BetaExplainer\cite{Explain, SubgraphX}. GNNExplainer and SubgraphX are some of the few edge explainers (out of four) applied to the first set of challenging datasets considered \cite{GraphXAI}. Another well-known method, PGExplainer\cite{PGExp}, generally underperforms GNNExplainer and SubgraphX. Thus, GNNExplainer and SubgraphX have both reasonable performance and the most comparable set of properties to BetaExplainer, suggesting they are well-positioned baselines for our study.
\subsection{Experimental Setup}

\subsubsection{Datasets}
Argawal et al. \cite{GraphXAI} propose that standardized methods to evaluate GNN explainability lack key characteristics. Limitations include few datasets with a notion of ground truth needed to measure explainer performance and under-represented real-world properties. To address these challenges, they developed the ShapeGGen simulator, which generates a variety of datasets with real-world properties and associated known ground truth. This simulator returns a diverse set of graph datasets given defined parameters. A house-shaped motif makes up the ground truth (i.e., all important edges) and generates 1200 subgraphs. Once generated, these subgraphs are connected so that each node has one or two ground truth motifs in its 1-hop neighborhood\footnote{A node's $l$-hop neighborhood is the set of nodes whose shortest path to the original node contains no more than $l$ edges}. The node's class of two is determined by the number of motifs in this neighborhood (zero if there is one motif and one if there are two). This structure makes up the first dataset with no challenging properties, SG-BASELINE, and provides the baseline for all remaining graphs.

The graphs with challenging properties are created by modifying one property of SG-BASELINE at once. Where the baseline graph is homophilic, in this case, indicating that nodes sharing an edge tended to contain the same number of motifs in their neighborhood and thus are of the same class, a graph can also be heterophilic. For the ShapeGGen simulator datasets, a heterophilic graph would be one where it is more likely that nodes with one motif in their 1-hop neighborhood will connect to nodes with two motifs in their 1-hop neighborhood. The second potentially challenging property is based on how correlated node class is to protected node features and how likely these features will be flipped. If not altered, it is assumed there is no correlation, and there is a 50\% likelihood the simulator would flip them. This setting ensures that the graph is entirely fair\footnote{Fairness measures the similarity of the model results on the data components deemed necessary to the model results on the original data/graph} since there is no added node feature bias which would affect the model but not actual node class \cite{GraphXAI}. Finally, the proportion of informative node features - or features correlated with the node class - to the total number of features may change where the baseline proportion of important features to total features is 4:11. The simulator may increase or decrease this ratio.

The resulting set of datasets examined is as follows, where each dataset alters only one property in relation to the baseline:

\begin{itemize}
  \item SG-BASE: A baseline dataset that is a large and homophilic graph with house ground-truth motifs, which is adapted to form the remaining datasets.
  \item SG-HETEROPHILIC: A modified version of the baseline dataset with a heterophilic graph.
  \item SG-UNFAIR: A modified version of the baseline dataset with a strongly unfair ground truth wherein protected node features are negatively correlated with the node class, and there is a 75\% chance node features will be flipped
  \item SG-MOREINFORM: A modified version of the baseline dataset with a high proportion of important to total features, 8:11.
  \item SG-LESSINFORM: A modified version of the baseline dataset with a low proportion of importance to total features, 4:21.
\end{itemize} 

While these datasets explore a variety of critical properties, the simulator does not return a sparse node feature dataset or dataset with a high proportion of zero node features with respect to the total number with known ground truth. As a result, there are no benchmarking results on how the explainer methods respond to this potentially challenging feature. Real-world datasets, particularly for biological applications, may have highly sparse features such as single-cell gene expression datasets \cite{SparseRNA}. This makes analyzing gene-gene interactions computationally, needed to ease wet lab analysis, more challenging \cite{GCNLongRange}. We hypothesize that sparse node features will influence explainer performance, which inspired us to include a second set of datasets: a baseline dataset at two different levels of sparsity. 

To examine the effects of sparsity on BetaExplainer, we needed a dataset with some ground truth gene-gene interaction graph. Otherwise, it would be unclear whether the method correctly selects important edges. We chose the SERGIO gene expression simulator \cite{SERGIO} as no ShapeGGen parameters allow us to control for this dataset explicitly. The SERGIO gene expression simulator is a widely used tool in the field of gene-gene interaction inference, capable of simulating the gene expression of a set of single cells using the chemical Langevin equation. This method allows the simulator to capture how regulator expression changes affect regulated gene expression, resulting in a gene regulatory network (GRN). Once SERGIO runs a set of simulations long enough to achieve steady-state, it samples gene expression from the graph to simulate a single-cell dataset. Specifically for our purposes, we use two cell types or classes, 100 genes per cell, and 1000 cells per cell type. Once sampled, we applied 25\% and 50\% random sparsity to the original dataset. This means we have two datasets with sparsity and known important edges - or the GRN that governs the cell expression (Fig. \ref{fig:Fig6}). Unlike the first set of datasets, these will both be graph classification problems: given a graph governing gene interactions and gene expression, a GNN predicts cell type.

However, in the real-world, the full groundtruth GRN may be unknown, suggesting simulating methods of approximating this underlying GRN are needed. We choose to create a graph based on genes that have a correlation of at least 0.35, indicating they are often expressed together, for each dataset. Due to limitations such as sparsity and regulation through intermediary genes, this graph contains a mix of true and false edges but is unable to capture all true gene-gene interactions while maintaining a computationally tractable graph. This allows us to approximate explainer accuracy on what it is given, though all explainers are unable to determine the importance of unseen edges.

\subsubsection{Implementation}
All trained GNN models used the Adam optimizer and cross-entropy loss, but model architecture (Fig. \ref{fig:Fig5}) and parameters (Table 1) vary to optimize the train and test accuracy for each dataset.

We initialize BetaExplainer with the node features and edge index of a given dataset, the $\alpha$ and $\beta$ parameters needed to initialize the Beta distribution, and the original model trained to classify outputs on the input data with epochs, learning rates, $\alpha$, and $\beta$ hyperparameters chosen based on the most balanced results across metrics (Table 2). Similarly, we chose the parameters resulting in the best performance for the GNNExplainer and the baselines (Table 2, Table 3).

\subsubsection{Metrics}
We judge whether BetaExplainer returned more important edges than GNNExplainer through accuracy metrics (specifically accuracy and F1 Scores to determine how well BetaExplainer returns important edges while ignoring unimportant edges) and unfaithfulness (to capture the similarity of model output on edges the explainer deems important to model output on all edges). For the accuracy metrics, calculation details varied depending on the dataset used. For the ShapeGGen \cite{GraphXAI} datasets, we focus on the best-performing subgraph since this was the method chosen for the simulator \cite{GraphXAI} and the whole graph for the SERGIO \cite{SERGIO} datasets for all analysis - qualitative (such as explanation graphs) and quantitative (the edge mask probability distributions and metrics). Letting $P$ denote precision and $R$ represent recall, we consider the F1 score which is computed as the following 
\begin{align}
\frac{2PR}{P + R}.
\end{align}
Accuracy was another area in which datasets differ. Given that $TP$ represents the number of true positives, $TN$ the true negatives, $FP$ the false positives, and $FN$ the false negatives, we used the traditional accuracy calculation for the SERGIO datasets:
\begin{align}
\frac{TP + TN}{TP + TN + FP + FN}.
\end{align}
We used two calculations for the number of false negatives: the first, containing all false negatives on the ground-truth elements in the input graph plus the missing ground-truth elements in the said graph, and the second, with just false negatives on the ground-truth elements for these datasets. We chose this method to capture explainer limitations as explainers may only analyze given data while ensuring the resulting metrics make sense. For the remaining datasets, to better mimic the ShapeGGen simulator \cite{GraphXAI}, we used the Jaccard Index:
\begin{align}
    \frac{TP}{TP + FP + FN + 1e-9}
\end{align}
While the $1e-9$ term is not strictly part of the Jaccard Index, the simulator incorporated it to avoid division by zero errors while maintaining rounding-accurate results, so we chose to include it as well \cite{GraphXAI}. The final metric calculated is unfaithfulness \cite{GraphXAI}, or
\begin{align}
1-\exp(-KL(f(X, G)||f(X, G_s)))
\end{align}
KL represents the KL divergence between the original GNN output given the full dataset versus the GNN output on the subgraph the explainer deems essential.

We calculated all metrics across ten random seeds, reporting the mean and standard error and ensuring randomness played less of a role in our results. We observed BetaExplainer's edge mask probability distribution of true and false positives over multiple runs with the same seed to best mimic the simulation structure and ensure randomness did not affect runs strongly. Visualizing the best-performing subgraph or graph over these runs for each explainer and the BetaExplainer edge mask probability distributions for true positives and true negatives also provided means to evaluate the model. To convey the uncertainty quantification that BetaExplainer produces, we also conveyed a graph weighing each edge based on probability modified through a variation of min-max scaling (supplementary) to clarify the range of probabilities taken.
\section{Results}
\subsection{BetaExplainer performs well on simulated datasets with challenging real-world properties}
\begin{figure}[H] 
\centering
\includegraphics[scale=0.75]{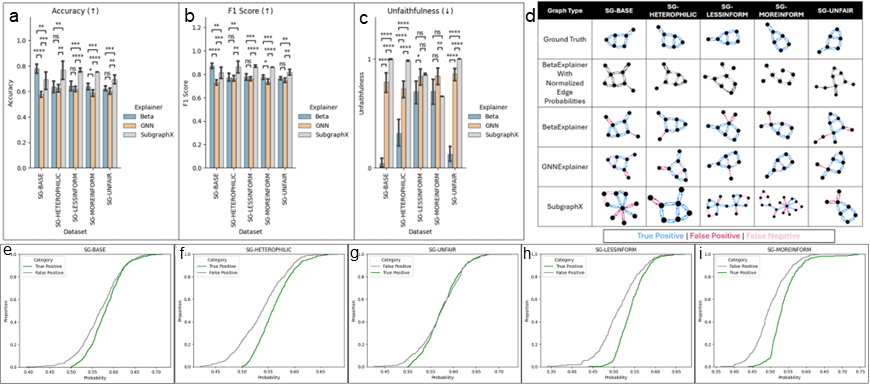}
\caption{We calculate the mean and standard errors of the Jaccard Index (a), F1 Score (b), and unfaithfulness (c) results and whether explainer differences are significant (ns: $0.05 < p \le 1$, *: $0.01 < p \le 0.05$, **: $0.001 < p \le 0.01$, ***: $0.0001 < p \le 0.001$, and ****: $p \le 0.0001$). We graph the best subgraphs for the datasets for each explainer versus the groundtruth (d), denoting true positive (blue), false positive (red), and false negative (pink) edges and weighting BetaExplainer edges by probabilities and empirical cumulative distribution (eCDF) of BetaExplainer probabilities for true and false positives with respect to the groundtruth (e, f, g, h, i)}
\label{fig:Fig3}
\end{figure}

We show that BetaExplainer achieves a better Jaccard Index, F1 Score, and unfaithfulness score than GNNExplainer on the five simulated datasets from \cite{GraphXAI}, with significant improvement for SG-BASE and SG-MOREINFORM for the first two metrics and across all datasets for the unfaithfulness score, and better unfaithfulness than SubgraphX on the SG-BASE, SG-HETEROPHILIC, SG-UNFAIR, and SG-LESSINFORM with significant improvements for the first three. We used the Mann-Whitney U test for $p$-value calculation (Fig. \ref{fig:Fig3} a-c) to test the significance of performance improvement. Explainers, particularly when faced with challenging properties such as heterophilic graphs, tend to generate unfaithful explanation graphs \cite{GraphXAI}, suggesting that these improvements are relevant. BetaExplainer optimizes the KL divergence between the GNN output on the masked graph and the original output, which is also measured by the unfaithfulness metric. This likely explains the decreased unfaithfulness score for BetaExplainer for most datasets and justifies the choice of our formulation. Problem formulation may account for SubgraphX achieving better unfaithfulness than BetaExplainer on the SG-MOREINFORM dataset. This dataset, in a way, is a baseline: it has even more informative features than even SG-BASELINE, suggesting the information provided by BetaExplainer's priors may make less of a difference. Furthermore, SG-MOREINFORM may not fully represent that challenges of real-world datasets, which are more likely to grapple with measurement errors.

Next, we comprehensively investigate BetaExplainer's performance in comparison to GNNExplainer and SubgraphX. We visualize the ground-truth sub-graphs and the explanation output for all the methods in Fig. \ref{fig:Fig3} d. We see that BetaExplainer generally returns more edges as important than GNNExplainer but balances the precision-recall trade-off well while maintaining a higher true positive rate than this baseline. SubgraphX seems to have a similar pattern as BetaExplainer but lacks the ranked scores for relevant edges. Since BetaExplainer is a probabilistic model, unlike SubgraphX, we can obtain the probability distribution of the edge mask scores. Figs. \ref{fig:Fig3} e-i plot the empirical cumulative distribution functions (eCDFs) of these probabilities for the true and false positive edges when compared to the ground truth. On average, we see that the probabilities assigned by BetaExplainer for true edges are higher than the ones for false edges. These probabilistic edge scores allow the user to select the most probable edges as the explanation based on a threshold of their choice. Such uncertainty quantification is not possible using other existing GNN interpretation methods (including GNNExplainer and SubgraphX), thus highlighting the need for probabilistic explanation models like BetaExplainer.

Considering that BetaExplainer improves upon GNNExplainer and SubgraphX in the unfaithfulness dimension for these simulations with challenging real-world properties, we will test it next for an additional graph simulation with sparse node features. This is relevant because many real-world datasets are sparse, such as scRNA-seq datasets, due to technical limitations \cite{SparseRNA}.

\subsection{BetaExplainer performs well on graph datasets with highly sparse node features.}
\begin{figure}[H] 
\centering
\includegraphics[scale=1]{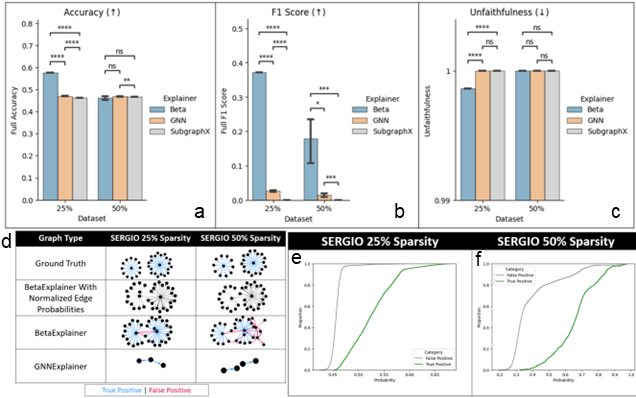}
\caption{We calculate the mean and standard errors of the accuracy (a), F1 score (b), and unfaithfulness (c) and the significant differences between explainer results (ns: $0.05 < p \le 1$, *: $0.01 < p \le 0.05$, **: $0.001 < p \le 0.01$, ***: $0.0001 < p \le 0.001$, and ****: $p \le 0.0001$). We graph best results per explainer for datasets (if true positives \textit{are} returned) (d), denoting true positive (blue), false positive (red), and false negative (pink) edges and weighting BetaExplainer edges by probabilities and the eCDF of the BetaExplainer probabilities for true and false positives with respect to the groundtruth (e, f)}
\label{fig:Fig4}
\end{figure}

\begin{figure}[!htbp] 
\centering
\includegraphics[scale=0.75]{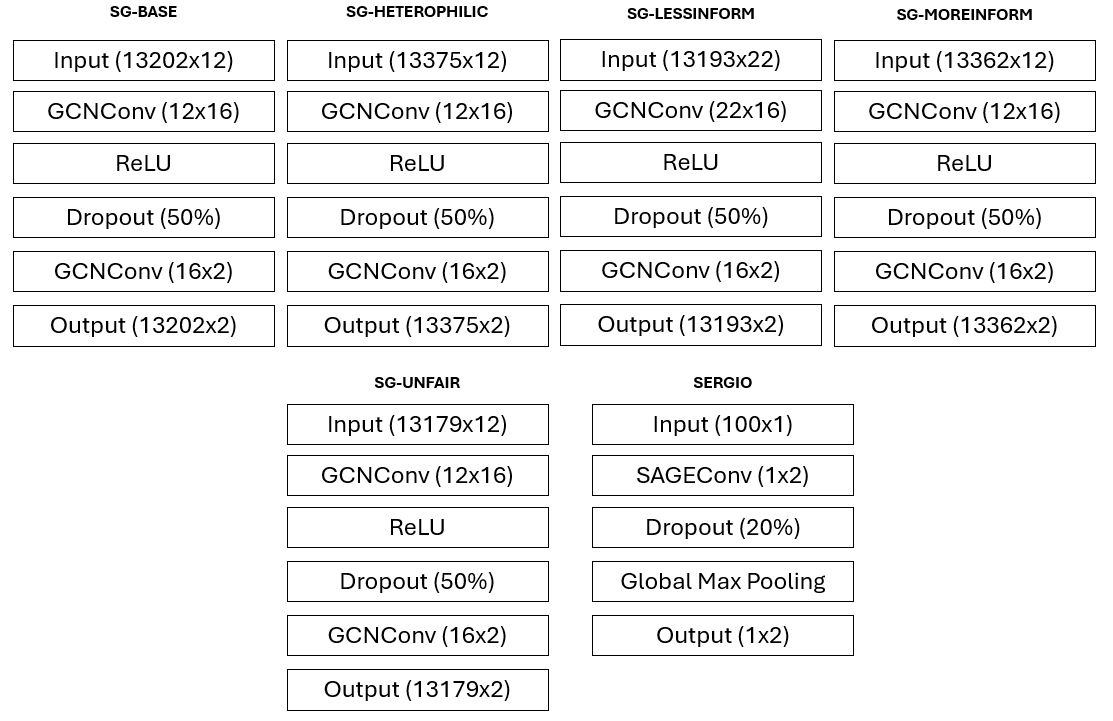}
\caption{The architecture for the input GNNs for the explainers are recorded for reproducibility}
\label{fig:Fig5}
\end{figure}

\begin{figure}[!htbp] 
\centering
\includegraphics[scale=0.5]{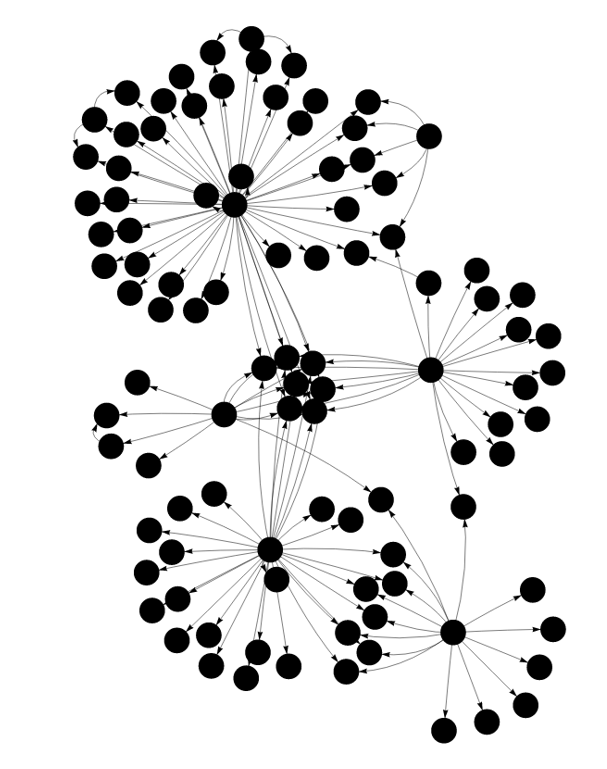}
\caption{This figure conveys the full ground-truth graph for all SERGIO datasets}
\label{fig:Fig6}
\end{figure}

\begin{figure}[!htbp]
\centering
\includegraphics[scale=0.75]{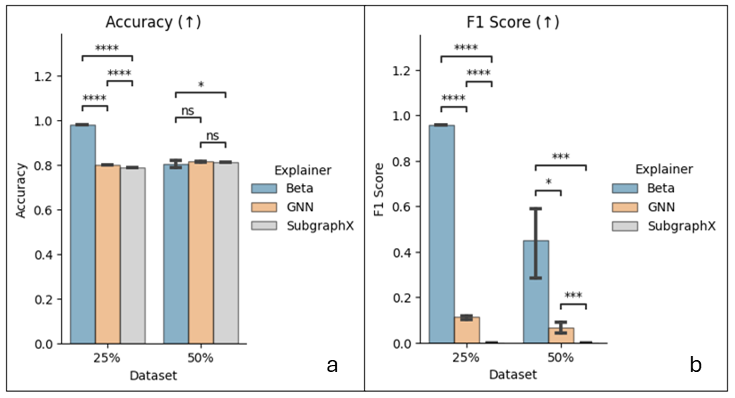}
\caption{When the false negatives from data collection are removed, the metrics increase as observed across a random set of seeds for accuracy (a) and F1 Score (b) with significant differences for F1 Score at both levels of sparsity (ns: $0.05 < p \le 1$, *: $0.01 < p \le 0.05$, **: $0.001 < p \le 0.01$, ***: $0.0001 < p \le 0.001$, and ****: $p \le 0.0001$)}
\label{fig:Fig7}
\end{figure}

BetaExplainer achieves similar accuracy as GNNExplainer and SubgraphX on the sparse SERGIO datasets and a significantly better F1 Score for both 25\% and 50\% sparse node feature datasets (Fig. \ref{fig:Fig4} a-b). These results again highlight the better precision-recall tradeoff of BetaExplainer than the baseline method. We expect overall low scores as the input graph that goes into the GNN calculated from correlation is sparse, to begin with, and this is standard in the field \cite{GRNConstruct}. We test this by excluding the false negatives representing the true edges absent from the correlation graph. This experiment confirms our assumption: we see a massive improvement in accuracy and F1 Scores (Fig. \ref{fig:Fig7} a-b). Furthermore, this calculation maintains the same pattern as the original -- BetaExplainer outperforms GNNExplainer and SubgraphX for the F1 Score metric (Fig. \ref{fig:Fig7} b).

Following the nuance of this dataset described above, the unfaithfulness metric (Fig. \ref{fig:Fig4} c) for this task is less reliable than the F1 score metric. The previous simulation datasets \cite{GraphXAI} contained all ground truth edges through the model training and explanation evaluation. However, it seems that using a graph with a different structure as input affects GNN training. This skews the unfaithfulness metric results for the explained graph since important model edges may not be in the ground truth graph, suggesting a tradeoff between accuracy and unfaithfulness metrics.

A qualitative analysis of the best-performing graphs for each explainer over each dataset confirms the primary driver of F1 Scores in Fig. \ref{fig:Fig4} d. While GNNExplainer may have higher precision, it comes at significant cost to the recall: it returns only two edges. The resulting SubgraphX graphs were not graphs as no edges were returned, indicating comparable accuracy due to accurate negative instances. BetaExplainer may have lower precision, but its recall is much higher. For real-world testing, researchers will need fewer false positive gene-gene interaction edges, suggesting in this case, BetaExplainer has a better precision-recall tradeoff balance.

Next, we examine the eCDFs of the true and false positive edges for BetaExplainer in Fig. \ref{fig:Fig4} e-f. Most true positives have a probability of 0.5 or greater, while most false negatives are less than or equal to the 0.5 bound. Users may prioritize a small set of the most essential edges in a real-world scenario from the high probabilistic scores obtained by BetaExplainer. One can be confident of this selection as this set contains very few false positives. This property is compelling as BetaExplainer can approximate the edge mask probabilistic distribution, prioritizing the actual important edges. 

Finally, BetaExplainer likely performs better than GNNExplainer due to the ability to capture the underlying distribution of edge importance by choosing the best $\alpha$ and $\beta$ parameters and thus is the better option for sparse datasets. Even if there is no improvement for non-sparse datasets, indicating either explainer is efficacious, improving upon challenging datasets is necessary. BetaExplainer improves upon critical metrics to the baseline across all challenging datasets tested, making it a helpful explanation method for the community. 
\section{Discussion}
BetaExplainer learns a probabilistic importance score for each edge by learning a Beta distribution. This is achieved by minimizing the KL divergence between the model output on the masked graph to the original output. By learning an importance score, users have a notion of uncertainty in edge importance. Furthermore, learning a probability distribution allows users to incorporate priors which provides the method with more information to adapt to datasets with challenging properties.

BetaExplainer achieves similar performance across accuracy and F1 scores to current state-of-the-art method GNNExplainer \cite{GNNExp} and SubgraphX \cite{SubgraphX}
for associated datasets and particularly achieves significantly better F1 Scores for the sparse node feature datasets. It also has similar if not better unfaithfulness results for almost all datasets, which are often significantly better, for the first five datasets \cite{GraphXAI}. Finally, BetaExplainer provides a measure of uncertainty, allowing users to focus most on most certain edges.

BetaExplainer has a few potential areas of improvement. It is sensitive to the number of GNN convolution layers due to the GNN oversmoothing issue. Addressing runtime is also a potential area to improve upon as BetaExplainer can take approximately 8.5 to 58 more seconds than GNNExplainer to run and from about 9.19 to 57.1 seconds longer than SubgraphX, and potentially more if batching is not used for graph datasets (Table 4). As BetaExplainer results are at minimum comparable, this may balance out results but finding ways to speed up the process is useful, especially as it is often able to outperform SubgraphX on the remaining datasets.

BetaExplainer model properties likely explain the improvements seen. We use a variety of $\alpha$ and $\beta$ parameters based on the datasets (Table 2), which likely perform best as they well-capture the underlying dataset properties. Exploration of these best-performing parameters for the dataset ensures a strong prior on important edges. Furthermore, BetaExplainer likely improves upon unfaithfulness results by optimizing KL divergence between the original model output on the full graph and model output on the masked graph.

In addition, results suggest users may apply BetaExplainer to a wide variety of datasets. Much like the SERGIO \cite{SERGIO} datasets, which represent gene expression data, many real-world expression datasets have sparse node features due to experimental limitations \cite{SparseRNA}. Since computational methods may clarify gene-gene interactions without needing expensive laboratory resources, BetaExplainer's ability to adapt to sparse datasets is critical \cite{GCNLongRange}. Graph classification problems are another area of exploration as this may prove to be a factor in the SERGIO results as they simulate a graph classification problem \cite{SERGIO}. Follow-up will determine whether sparsity remains the only factor. In addition, graphs representing protein structure are heterophilic, since different node classes represent different amino acid types \cite{GNNHetSurvey}. The improved unfaithfulness on the simulated heterophilic dataset may suggest BetaExplainer will continue to perform well in this area. Further exploration of these datasets will be critical to understand BetaExplainer's potential and important elements of real-world graphs.

\section*{Code availability}

All code is available under the open-source MIT license at \url{https://github.com/wsloneker/BetaExplainerDemo}.

\section*{Acknowledgments} 

We are grateful to Ghulam Murtaza for helping us understand challenges with the SERGIO datasets and for aid in revising, Michal Golanvesky for providing seed resources, and Alexandra Miller for aid in proofreading. This research was conducted using computational resources and services provided by the Center for Computation and Visualization at Brown University.

\section*{Funding}

This research was also supported in part by a David \& Lucile Packard Fellowship for Science and Engineering awarded to LC. Any opinions, findings, conclusions, or recommendations expressed in this material are those of the author(s) and do not necessarily reflect the views of any of the funders.

\section*{Competing interests}

LC is an employee of Microsoft and owns equity in the company. All other authors have declared that they have no competing interests.

\newpage
\section{Supplementary}

\textbf{Table 1.} The hyperparameters for the trained GNNs for the explainers are conveyed

\begin{center}
\begin{tabular}{||c c c c c||} 
 \hline
 \textbf{Dataset} & \textbf{Learning Rate} & \textbf{Weight Decay} & \textbf{Seed} & \textbf{Epochs} \\ [0.5ex] 
 \hline\hline
  \hline
 SG-BASE & 0.16 & 0.0001 & 1 & 2000\\
 \hline
  SG-HETEROPHILIC & 0.1 & 5e-5 & 10 & 2000\\
 \hline
   SG-UNFAIR & 0.15 & 0.0001 & 4 & 2000\\
 \hline
   SG-LESSINFORM & 0.05 & 0.001 & 1000 & 2000\\
 \hline
   SG-MOREINFORM & 0.05 & 0.001 & 400 & 2000\\
 \hline
   SERGIO 25\% Sparsity & 0.001 & NA & 200 & 50\\
 \hline
    SERGIO 50\% Sparsity & 0.001 & NA & 200 & 50\\
 \hline
\end{tabular}
\end{center}

\textbf{Table 2.} The hyperparameters for GNNExplainer and BetaExplainer are conveyed

\begin{center}
\begin{tabular}{||c c c c c c c c||} 
 \hline
 \textbf{Dataset} & \textbf{Explainer} & \textbf{Learning Rate} & \textbf{Alpha} & \textbf{Beta} & \textbf{Batch} & \textbf{Epochs} & \textbf{Type} \\ [0.5ex] 
 \hline\hline
  \hline
 SG-BASE & Beta & 0.05 & 0.8 & 0.6 & NA & 25 & NA \\
 SG-BASE & GNN & 1e-5 & NA & NA & NA & 200 & NA \\
 \hline
  SG-HETEROPHILIC & Beta & 0.05 & 0.7 & 0.6 & NA & 25 & NA  \\
 SG-HETEORPHILIC & GNN & 1e-5 & NA & NA & NA & 200 & NA \\
 \hline
  SG-UNFAIR & Beta & 0.05 & 0.8 & 0.6 & NA & 25 & NA \\
 SG-UNFAIR & GNN & 1e-5 & NA & NA & NA & 200 & NA \\
 \hline
  SG-LESSINFORM & Beta & 0.05 & 0.6 & 0.6 & NA & 25 & NA \\
 SG-LESSINFORM & GNN & 1e-5 & NA & NA & NA & 200 & NA  \\
 \hline
  SG-MOREINFORM & Beta & 0.05 & 0.8 & 0.8 & NA & 25 & NA  \\
 SG-MOREINFORM & GNN & 1e-5 & NA & NA & NA & 200 & NA \\
 \hline
  SERGIO 25\% Sparsity & Beta & 0.001 & 0.55 & 0.65 & NA & 25 & NA \\
 SERGIO 25\% Sparsity & GNN & 0.00001 & NA & NA & 300 & 300 & Phenomenon \\
 \hline
  SERGIO 50\% Sparsity & Beta & 0.01 & 0.5 & 0.95 & NA & 25 & NA \\
 SERGIO 50\% Sparsity & GNN & 0.0001 & NA & NA & 859 & 300 & Phenomenon \\
 \hline
\end{tabular}
\end{center}

\textbf{Table 3.} The hyperparameters for SubgraphX are conveyed
\begin{center}
\begin{tabular}{||c c c c c c c||} 
 \hline
  \textbf{Dataset} & \textbf{Rollout} & \textbf{Min Atoms} & \textbf{C Puct} & \textbf{Expand Atoms} & \textbf{Sample Num} & \textbf{Node} \\ [0.25ex] 
 \hline\hline
 \hline
 SG-BASE & 5 & 8 & 4.763396105210084 & 16 & 10 & 8966 \\
 \hline
 SG-HETEROPHILIC & 6 & 10 & 7.948316238894563 & 6 & 3 & 7324 \\
 \hline
 SG-UNFAIR & 8 & 7 & 6.933836891280444 & 8 & 8 & 3046 \\
 \hline
 SG-LESSINFORM & 5 & 4 & 4.919019371456805 & 4 & 3 & 2913 \\
 \hline
 SG-MOREINFORM & 7 & 9 & 7.318672180500459 & 3 & 10 & 3403 \\
 \hline
  SERGIO 25\% Sparsity & 20 & 1 & 8.165129763712843 & 5 & 6 & n/a \\
 \hline
  SERGIO 50\% Sparsity & 4 & 1 & 2.0727783316948987 & 7 & 9 & n/a \\
\hline
\end{tabular}
\end{center}

To visualize the uncertainty quantification of images, a variation of min-max scaling was used to differentiate weights. This was achieved by using a variation of max-min scaling as follows to ensure all values were greater than zero with $M$ representing the maximum value of all probabilities for accepted edges (i. e. at least the minimum bound); $m$, the minimum of all these probabilities, and $p$ the probability used as input:

\begin{align}
    \frac{p - m + 1e-5}{M - m}
\end{align}

\textbf{Table 4.} The average runtime (in seconds) over 50 epochs was calculated over 25 runs, and the average of the average runtime per epoch was calculated
\begin{center}
\begin{tabular}{||c c c c||} 
 \hline
 \textbf{Dataset} & \textbf{Explainer} & \textbf{Full Runtime (s)} & \textbf{Average Epoch Runtime (s)} \\ [0.5ex] 
 \hline\hline
  \hline
 SG-BASE & GNN & \textbf{2.09} & \textbf{4.19e-2} \\ 
 SG-BASE & SubgraphX & 20.1 & n/a \\ 
 SG-BASE & Beta & 11.2 & 2.24e-1 \\
 \hline
  SG-HETEROPHILIC & GNN & 3.35 & \textbf{6.69e-2} \\ 
  SG-HETEROPHILIC & SubgraphX & \textbf{0.464} & n/a \\ 
 SG-HETEROPHILIC & Beta & 15.4 & 3.08e-1 \\
 \hline
  SG-LESSINFORM & GNN & 2.96 & \textbf{5.93e-2} \\ 
  SG-LESSINFORM & SubgraphX & \textbf{2.21} & n/a \\ 
 SG-LESSINFORM & Beta & 11.4 & 2.29e-1 \\
 \hline
  SG-MOREINFORM & GNN & 2.07 & \textbf{4.14e-2} \\ 
  SG-MOREINFORM & SubgraphX & \textbf{0.605} & n/a \\ 
 SG-MOREINFORM & Beta & 14.1 & 2.82e-1 \\
 \hline
  SG-UNFAIR & GNN & 2.86 & \textbf{5.72e-2} \\
  SG-UNFAIR & SubgraphX &  \textbf{2.71} & n/a \\ 
 SG-UNFAIR & Beta & 21.8 & 4.35e-1 \\
  \hline
 SERGIO 25\% Sparsity & GNN & 0.127 & \textbf{5.10e-3} \\ 
   SERGIO 25\% Sparsity & SubgraphX & \textbf{9.32e-2} & n/a \\ 
 SERGIO 25\% Sparsity & Beta & 515 & 10.3 \\
 SERGIO 25\% Sparsity With Batching & Beta & 57.2 & 1.14 \\
 \hline
  SERGIO 50\% Sparsity & GNN & \textbf{3.19e-2} & \textbf{1.27e-3} \\ 
  SERGIO 50\% Sparsity & SubgraphX & 0.208 & n/a \\ 
 SERGIO 50\% Sparsity & Beta & 415 & 8.3 \\
 SERGIO 50\% Sparsity With Batching & Beta & 57.2 & 1.14 \\
 \hline
\end{tabular}
\end{center}

\end{document}